\documentclass{bmvc2k}
\usepackage{graphicx}
\usepackage{amsmath,amssymb}
\usepackage{algorithm}
\usepackage{algpseudocode}

\title{ViCE: Improving Dense Representation Learning by Superpixelization and Contrasting Cluster Assignment}

\addauthor{Robin Karlsson}{karlsson.robin@g.sp.m.is.nagoya-u.ac.jp}{1}
\addauthor{Tomoki Hayashi}{hayashi.tomoki@g.sp.m.is.nagoya-u.ac.jp}{1}
\addauthor{Keisuke Fujii}{fujii@i.nagoya-u.ac.jp}{1}
\addauthor{Alexander Carballo}{alexander@g.sp.m.is.nagoya-u.ac.jp}{1}
\addauthor{Kento Ohtani}{ohtani.kento@g.sp.m.is.nagoya-u.ac.jp}{1}
\addauthor{Kazuya Takeda}{kazuya.takeda@nagoya-u.jp}{1,2}

\addinstitution{
 Graduate School of Informatics\\
 Nagoya University\\
 Aichi, Japan
}
\addinstitution{
 Tier IV Inc.\\
 Tokyo, Japan
}

\runninghead{R. Karlsson et al.}{ViCE: Improving dense representation learning}

\def\etal{\emph{et al}\bmvaOneDot}

\begin{document}

\maketitle

\begin{abstract}
Recent self-supervised models have demonstrated equal or better performance than supervised methods, opening for AI systems to learn visual representations from practically unlimited data. However, these methods are typically classification-based and thus ineffective for learning high-resolution feature maps that preserve precise spatial information. This work introduces superpixels to improve self-supervised learning of dense semantically rich visual concept embeddings. Decomposing images into a small set of visually coherent regions reduces the computational complexity by $\mathcal{O}(1000)$ while preserving detail. We experimentally show that contrasting over regions improves the effectiveness of contrastive learning methods, extends their applicability to high-resolution images, improves overclustering performance, superpixels are better than grids, and regional masking improves performance. The expressiveness of our dense embeddings is demonstrated by improving the SOTA unsupervised semantic segmentation benchmark on Cityscapes, and for convolutional models on COCO. Code is available at \url{https://github.com/robin-karlsson0/vice}.
\end{abstract}


\section{Introduction}
\label{sec:introduction}

Progress in general computer vision tasks in the past decade has been based on supervised learning with large datasets annotated by human labelers~\cite{Krizhevsky2012ImageNetConv}. Arguments are made that generalizable and robust computer vision models have not yet been achieved, and further increasing the amount of labeled data is unsustainable~\cite{LeCunn2020AAAI20SSL, Marcus2019RebootingAI}. One hypothesis is that learning

\begin{figure}[h]
\begin{center}
\includegraphics[width=1.0\textwidth]{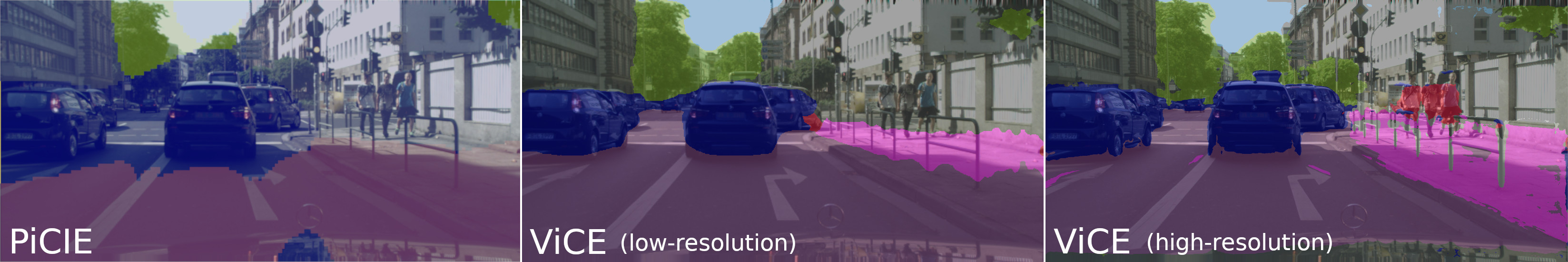}
\end{center}
   \caption{ViCE learns dense semantic embeddings from raw image data. Unsupervised semantic segmentation experiments show that our embedding maps are semantically richer and fit the content better compared to the SOTA baseline PiCIE~\cite{Cho2021PiCIE}. Superpixelization further improves our results by enabling dense contrastive learning over high-resolution images.}
\label{fig:front_cover}
\end{figure}

\noindent
from top-down categorization (``what it \textit{is}") from semantically vague and inconsistent human annotation could be a limiting factor~\cite{Efros2021SSLBottomUP}. Instead, cognitive science tells us that learning from bottom-up association (``what it is \textit{like}") may be more similar to how visual concepts emerge for humans~\cite{Rosch1973NatCat,Medin1978ContextTheory,Nosofsky1986AttSimIdCatRel,Nosofsky1992ExemplarConnectionistLearning}. The success of bottom-up learning for word embeddings in natural language processing (NLP)~\cite{Harris1954DistributionalHypothesis,Mikolov2013EffWordRep,Mikolov2013DistributedRO} further strengthens the hypothesis. Recent self-supervised computer vision methods show promise in this direction with results approaching or even surpassing those of supervised methods~\cite{Goyal2021SEER}. However, these methods are classification-based and thus ineffective for learning high-resolution dense feature maps. Such maps are needed to associate semantic embeddings to spatial regions in vision inputs.

We introduce a method for improving the effectiveness of self-supervised classification methods for dense representation learning by decomposing images into a small set of visually coherent regions using superpixelization. We demonstrate how applying the method enables the contrasting cluster assignments method SwAV~\cite{Caron2020SwAV} to learn dense representations.
The contributions of our paper are as follows:
\begin{itemize}
    \setlength\itemsep{-0.5em}
    \item{A new conceptual approach to represent high-resolution images as semantically rich embedding maps partitioned into distinct, coherent regions, represented by a latent \textbf{Vi}sual \textbf{C}oncept \textbf{E}mbedding (ViCE), analogous to word embeddings in NLP.}
    \item {Introduce superpixelization as a natural hierarchical region decomposition for dense contrastive learning in unsupervised semantic segmentation of high-resolution images. We demonstrate how to effectively implement self-supervised classification methods with region decomposition.}
    \item{Present SOTA unsupervised semantic segmentation results on Cityscapes, and for convolutional models on COCO.}
    \item{Experimentally demonstrate; Online contrasting cluster assignment~\cite{Caron2020SwAV} improves dense representation learning performance compared with offline clustering~\cite{Caron2018DeepCluster, Cho2021PiCIE}. Image decomposition by superpixelization improves performance, reduces computational time, and is more effective than grids. The ability to use high-resolution images improves performance. Contextual region masking improves performance.}
    
\end{itemize}

\section{Related work}
\label{sec:related_work}

\textbf{Self-supervised visual representation learning} Early works experimented with pretext tasks as a substitute for human annotations~\cite{Doersch2015PatchPred,Pathak2016Inpainting, Noroozi2016Jigsaw,Bucci2021SSLAcrossDomains,Gidaris2018RotPred,Zhang2016ColorfulIC}. Recent work demonstrates that image-level embedding classification with cross-entropy minimization on large datasets is a more effective approach capable of surpassing supervised pretraining~\cite{Goyal2021SEER, Caron2021DINO}. Contrastive methods~\cite{Chopra2005TripletLoss, Chen2020SimCLR, Oord2018ContrastivePredCoding, He2020MoCo} learn discriminative latent embedding vectors for images by ``pulling together" views of the same image, and ``pushing away" embeddings of different images. Recent non-contrastive methods~\cite{Zbontar2021BarlowTwins, Grill2020BYOL, Caron2021DINO} demonstrate approaches to avoid negative sampling to improve computational efficiency. Clustering methods \cite{Caron2018DeepCluster, Asano2020SeLA,Caron2020SwAV, Caron2019Noncurrated, Zhan2020OnlineDC, Yang2019DeepClusterGMVA, Li2021ContClust} simultaneously discovers a set of clusters or prototypes, and learns discriminative image embeddings. Contrary to contastive methods, the objective does not have to be approximated as optimizing over the entire set of negative representative clusters is tractable. DeepCluster~\cite{Caron2018DeepCluster} iteratively performs K-means clustering over the entire dataset and learns an embedding model and classification head to predict the cluster assignment. SeLA~\cite{Asano2020SeLA} presents a principled formulation for clustering and representation learning as a single optimization objective, by casting cluster assignment as an optimal transport problem~\cite{Levy2018OptimalTransport, Cuturi2013OptimalTransport}. SwAV~\cite{Caron2020SwAV} and ODC~\cite{Zhan2020OnlineDC} demonstrate that clustering can be done online per batch to increase learning efficiency.

\noindent
\textbf{Dense representation learning} Recent clustering-based methods approach dense representation learning as an instance segmentation problem~\cite{Chen2021MultiSiamSM, Li2021DenseSC, Henaff2021EfficientVP, Zhang2020HierarchicalGrouping} and regional feature correspondence~\cite{Wang2012DenseCL, Li2021DSCL, Xie2021PixPro}. These methods are purposed for pretraining backbones and generally output small feature maps (e.g. 7x7), in contrast to our method. Similarly to our method, VADeR~\cite{Pinheiro2020VADeR} learns dense representations by contrasting pixel-level embeddings in augmented views. Our method improves on VADeR by allowing training on larger feature maps (512x512 vs. 56x56 px), more views, optimization without a negative sample memory bank, and contextual region masking. Self-supervised object detection~\cite{Bar2021DETReg, Wei2021AligningPF, Yang2021InstanceLF, Dai2021UPDET, Wang2021DenseCL, Xiao2021RegionSR} learns expressive embeddings for plausible object proposal regions sampled randomly or heuristically~\cite{Uijlings2013SelectiveSearch}. Masked image modeling (MIM)~\cite{He2021MaskedAA, Xie2021SimMIMAS, Bao2022BEiT, Chen2022CAE} demonstrates strong representation learning capability surpassing contrasting views. However, all these models output low-resolution feature maps. In contrast, our method ViCE generates precise object-fitting semantic partitioning even for high-resolution images.

\noindent
\textbf{Unsupervised semantic segmentation} Existing works leverage self-supervised clustering approaches to learn coherent semantic groupings from mutual information~\cite{Ji2019IIC, Ouali2020AutoregSem}, geometric equivariance~\cite{Cho2021PiCIE}, and GAN-based approaches~\cite{Chen2019UnsupervisedOS,Bielski2019EmergenceOO}. Other works~\cite{Hoyer2021ThreeWT,Vu2019DADA} leverages self-supervised depth map estimation~\cite{Godard2019DiggingIS,Bello2020ForgetAT} for enhancing semantic segmentation performance. Recently, DINO~\cite{Caron2021DINO} demonstrated that attention maps for semantic objects naturally emerge for self-supervised Vision Transformer (ViT) models~\cite{Dosovitskiy2021ViT, Vaswani2017AttentionIA}. STEGO~\cite{Hamilton2022STEGO} presents a method to distill features from DINO and achieve SOTA results. Our work improves learning efficiency also on high-resolution images by contrasting cluster assignment over superpixels.

\noindent
\textbf{Image decomposition by superpixeliation} Prior work which visually groups pixels includes semi- and weakly supervised models~\cite{Franchi2021RobustSS, Yi2022WeaklysupervisedSSGuided, Kwak2017WeaklySSPooling}, and methods bootstrapping from pretrained saliency~\cite{VanGansbeke2021UnsupervisedSS} and contour detector~\cite{Zhang2020UnsupHiearchicalGrouping, Hwang2019SegSort, Ke2022UnsupervisedHS} models. We utilize visual grouping without depending on pretraining and not only as an inductive bias, but to perform contrastive learning over a set of visually coherent regions instead of individually meaningless pixels. Ouyang~\etal~\cite{Ouyang2020SelfSupSuperpixels} uses self-supervised learning to map superpixel regions between augmented views for transferring semantic labels in annotated samples to corresponding regions in unannotated samples. \cite{Kanezaki2018UnsupSegSuperpixels, Mirsadeghi2021InMARS} uses superpixels to refine the unsupervised segmentation output. In contrast, our method uses superpixels to learn semantics from high-resolution images without annotated data.

\section{ViCE: Visual Concept Embeddings}
\label{sec:vice}

The concept of ``the thing in itself" in Kantian philosophy denotes the existence of objects as they are independent of observation. Similarly, one can view natural images perceived by a photometric sensor to be generated from a set of latent semantic visual concepts. We model this process by a model $f(X | Z)$ that generates the observable pixel appearance $X$ of semantic entities represented by a set of latent visual concepts $C = (c^{(1)}, \dots, c^{(K)})$, encoded into a dense embedding map $Z$. Our method is based on learning a function $f_\theta$ to approximate the inverse mapping $f^{-1}(Z | X)$ while simultaneously discovering the set of latent visual concepts $C$. The problem of finding the inverse mapping is called vision as inverse graphics~\cite{Kersten2004PerceptionBayesianInf, Kersten2006AnalysisSynthesis, Comellas2019VisionInvGraphics}. We propose to learn a mapping $f_\theta$ that predicts the same visual concept embedding map $Z \in \mathbb{R}^{D \times H \times W}$ with the same spatial resolution as the input image $X \in \mathbb{R}^{3 \times H \times W}$ for all mutually co-occurring abstract pixel patterns generated from augmented views $\tilde{X}^{(m)}$. All views contain one subregion representing the same content, but with different pixel appearances and surrounding context.
\begin{equation}
\label{eq:inverse_mapping}
   f_\theta ( \tilde{X}^{(m)} ) \simeq Z \quad \forall m \in (1, \dots, M)
\end{equation}
\begin{figure}
\begin{center}
\includegraphics[width=1.0\textwidth]{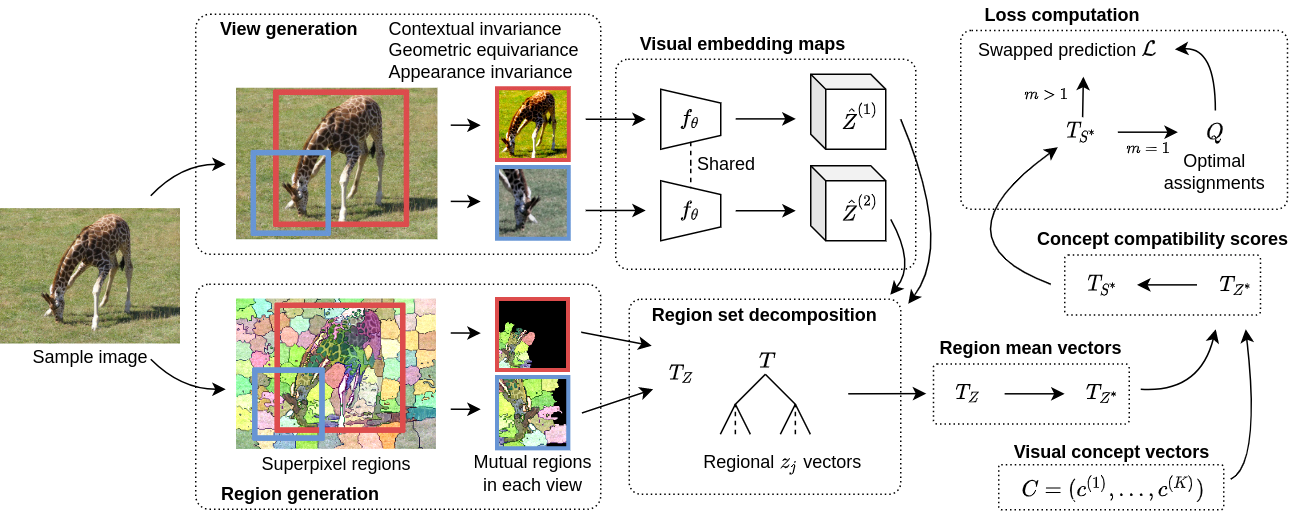}
\end{center}
   \caption{Overview of ViCE. A training iteration starts by generating $M$ augmented views. First, we partition the image into $I$ mutually common superpixel regions. The model $f_\theta$ transforms view images into visual concept embedding maps $\hat{Z}^{(m)}$. All vectors $z_j$ are arranged in a tree structure $T_Z$ used to conveniently organize indices of corresponding regions. A mean vector $z^*_i$ is computed for each region. Next, we score each $z^*_i$ in terms of closeness to each concept vector $c^{(k)}$, resulting in region-specific score vectors $s^*_i$.}
\label{fig:overview}
\end{figure}
We relate our approach to discovering semantic meanings for pixels to discovering semantic meanings for words in NLP similar to recent MIM works~\cite{Bao2022BEiT, Zhou2021iBOT, Chen2022CAE, Xie2021SimMIMAS}. Methods to learn semantically rich word embeddings~\cite{Mikolov2013EffWordRep,Mikolov2013DistributedRO,Pennington2014GloVe} are based on
co-occurrence~\cite{Harris1954DistributionalHypothesis} and context~\cite{Devlin2019BERT,Peters2018ELMO} of individually meaningless tokens. Each visual concept vector $c$ corresponds to a distinct visual concept primitive or basis vector, and visual concepts are linear combinations of these primitives. The set of concepts $C$ is known and finite, ensuring tractable probabilistic enumeration over possible configuration akin to successful probabilistic language modeling approaches in NLP~\cite{Devlin2019BERT, Radford2018GPT}. We choose to demonstrate our method with the recent SOTA self-supervised learning method SwAV~\cite{Caron2020SwAV} to learn both $f_{\theta}$ and $C$, though in principle any cluster-based self-supervised method can be used. Fig.~\ref{fig:overview} shows an overview of our method.

\subsection{Decomposing images into visually coherent regions}
A high-resolution image contains millions of individually meaningless and mostly redundant pixels. However, it is known that training on high-resolution images is beneficial for learning to segment small objects such as poles and pedestrians~\cite{Chen2018DeepLabV2}. Nevertheless, naively applying self-supervised representation learning methods based on vector comparison on high-resolution embedding maps is inefficient. To solve this problem, we propose to decompose the image into a small set of visually coherent regions using superpixelization~\cite{Ren2003Superpixels} and apply representation learning methods to this greatly reduced set of elements.
Superpixel methods like Simple Linear Iterative Clustering (SLIC)~\cite{Achanta2010SLICSuperpixels} reduce elements by $\mathcal{O}(1000)$, transforming an image from millions of pixels into less than a thousand regions. We choose SLIC because of advantages~\cite{AchantaSLICSOTA} such as more uniform region distribution compared to graph-based methods~\cite{Pedro2004FHsuperpixels}. 
In contrast to grid decomposition, which is the standard for ViT models~\cite{Dosovitskiy2021ViT, Caron2021DINO}, superpixels can preserve detail by representing thin and small patches like poles as distinct regions while requiring 75\% fewer elements on average with the same base element size.
While in this paper our objective is to show that even the simplest form of region decomposition is useful, it is likely that leveraging learning-based superpixelization methods~\cite{Arbelaez2014MGC, Locatello2020SlotAttention, Weinzaepfel2022Superfeatures} can further improve performance.

\subsection{View generation and contextual region masking}
\label{sec:view_generation}
We generate augmented views for discerning the latent semantic visual concepts through photometric invariance~\cite{Chen2020SimCLR}, and geometric equivariance~\cite{Cho2021PiCIE}. We introduce region masking as an additional augmentation for contextual invariance shown to improve performance.
To generate views with different contexts, we first sample a center point $(x, y)^*$ in the image. Sampling is done in content-rich regions to better satisfy the equipartitioning of concepts assumption~\cite{Asano2020SeLA, Caron2020SwAV} for each training batch. We found that probabilistic sampling from a Gaussian filtered Canny edge detection map\cite{Canny1986CannyEdge} is a useful measure of image content. Views $\tilde{X}^{(m)}$ are generated by sampling $M$ view centers $(x, y)^{(m)}$ around $(x, y)^*$ while ensuring a mutual image subregion exists.
We generate geometrically equivariant views by first sampling a resize coefficient $\beta^{(m)}$ for each view $m$. $\beta$ determines the size of the cropped view region as exemplified by the red and blue crop regions in Fig.~\ref{fig:overview}. All view crops are resized to the common view size, thus enforcing the model to learn resolution invariant representations. All views are randomly flipped horizontally.
All views are augmented by random color distortion and Gaussian blurring before normalization to learn appearance invariant visual concepts~\cite{Chen2020SimCLR, Xiao2021WhatShouldNot, Wen2021TowardUnderstanding}. A ratio of superpixel regions is masked with noise as a means to learn robust features and alleviate the shortcut learning problem~\cite{Geirhos2020ShortcutLearning}. We provide the view generation algorithm as pseudocode in the Appendix.

\subsection{Learning algorithm}

The objective $\mathcal{L}_{cl}$ is designed to simultaneously learn the mapping function $f_\theta$ in Eq.~\ref{eq:inverse_mapping}, and optimize the distribution of latent visual concepts $C$. The algorithm can be viewed as an extension of SwAV~\cite{Caron2020SwAV} to the problem of learning dense embedding maps. We refer to prior work for an explanation of SwAV~\cite{Caron2020SwAV, Asano2020SeLA, Cuturi2013OptimalTransport, Kaiser2021UnderstandingSwAV}. The rest of this section explains the flow of a training iteration as visualized in Fig.~\ref{fig:overview}. We provide pseudocodes in the Appendix.

A training iteration starts by partitioning an image $X^{(n)} \in \mathbb{R}^{3 \times H \times W}$ with height $H$ and width $W$ into a superpixel region map $A^{(n)} \in \mathbb{R}^{H \times W}$, with integer values specifying every pixel’s region index. Next, a set of $M$ augmented views $\tilde{X}^{(n)} = \{ \tilde{X}^{(1,n)}, \dots, \tilde{X}^{(M,n)} \}$ and corresponding superpixel map crops $\tilde{A}^{(n)} = \{ \tilde{A}^{(1,n)}, \dots, \tilde{A}^{(M,n)} \}$ of size $h$ and $w$ are generated for each image as explained in Sec.~\ref{sec:view_generation}. $\tilde{A}^{(n)}$ is processed to contain only mutual regions existing in all views. 
The learned function $f_\theta$ transforms $\tilde{X}^{(n)}$ into a normalized visual embedding tensor $\hat{Z}^{(n)} \in \mathbb{R}^{D \times h \times w}$. Next $\hat{Z}^{(n)}$ is decomposed region-wise into row vectors $z_j \in \mathbb{R}^D$ and stored in a tree structure $T_Z$ used to conveniently organize indices of corresponding regions $i$ in view $m$ of image $n$. Vectors of non-mutual regions are discarded. A single mean vector $z^{(i,m,n)*}$ is computed to represent each region $i$ and stored in $T_{Z^*}$. Each vector $z^{(i,m,n)*}$ is scored in terms of compatibility or closeness to each visual concept vector $C = (c^{(1)}, \dots, c^{(K)})$ by computing the following matrix product
\begin{equation}
    s^* = (z^*)^T C
\end{equation}
\noindent with $C \in \mathbb{R}^{D \times K}$ represented as an optimizable weight matrix. Note that the dot product $z \cdot c$ equals the cosine distance as both vectors are normalized. All regional score vectors $s^{(i,m,n)*}$ are stored in a tree structure $T_{S^*}$. The concept assignments $q^{(i)}$ are determined by optimally distributing $s^{(i,m,n)*}$ uniformly over all concepts $c^{(k)}$ so that the overall compatibility between all $s^{(i)}$ and $c^{(k)}$ are maximized for regions in the primary view $m=1$~\cite{Caron2020SwAV}. We compute $q^{(i)}$ efficiently by the Sinkhorn-Knopp algorithm~\cite{Asano2020SeLA, Cuturi2013OptimalTransport}. A FIFO queue of accumulated $s^{(i,1,n)*}$ vectors is used to improve the empirical approximation of a uniform distribution of concepts~\cite{Asano2020SeLA, Caron2020SwAV}. The swapped prediction learning objective~\cite{Caron2020SwAV} is
\begin{equation}
\label{eq:clustering_loss}
  \mathcal{L}_{cl} = -\frac{1}{N (M-1)} \sum^N_{n = 1} \sum^M_{m = 2} \frac{1}{I} \sum^I_{i = 1} q^{(i)} log \: \sigma \left( \tfrac{1}{\tau} s^{(i,m)*} \right)
\end{equation}
\noindent where $\sigma ()$ is the softmax function and $\tau$ is temperature. Two normalized embeddings $z^{(a)}$ and $z^{(b)}$ are compared for semantic similarity using the dot product. This operation is equivalent to comparing two word embeddings by cosine distance~\cite{Mikolov2013DistributedRO, Mikolov2013EffWordRep}.

\setlength{\tabcolsep}{4pt}
\begin{table}[ht]
\begin{center}
\caption{Representation quality experiment results on low- and high-resolution images.}
\begin{tabular}{cccc|cccc}
\hline
Model &        & mIoU  & Acc. & Model  &       & mIoU  & Acc.  \\
\hline
\multicolumn{4}{c|}{\textit{COCO}} & \multicolumn{4}{c}{\textit{Cityscapes}}\\
ResNet50~\cite{He2016ResNet50}          & C 27 & 8.9  & 24.60 &   ResNet50~\cite{He2016ResNet50} &  C 27   & -  & - \\
MoCoV2~\cite{Chen2020MoCoV2}          &  C 27 & 10.40  & 9.60 &     MoCoV2~\cite{Chen2020MoCoV2} &   C 27   & -  & - \\
DINO$^*$~\cite{Caron2021DINO}          & C 27 & 9.60  & 30.50 &  DINO$^*$~\cite{Caron2021DINO}   & C 27  & -  & - \\
IIC~\cite{Ji2019IIC}          & C 27 & 6.71  & 21.79 & IIC~\cite{Ji2019IIC}  &     C 27.     & 6.35  & 47.88 \\
PiCIE~\cite{Cho2021PiCIE} & C 27   & 13.84 & 48.09 & PiCIE~\cite{Cho2021PiCIE} & C 27   & 12.31 & 65.50 \\
  & C 27$^\diamond$   & 14.60 & 48.37 &   & C 27$^\diamond$  & 11.85 & 64.29 \\
  & C 27$^\star$   & 9.27 & 38.31 &   & C 27$^\star$  & 8.80 & 82.48 \\
 & C 128$^\star$   & 10.75 & 49.81 &  & C 128$^\star$   & 7.97 & 56.52 \\
 & C 256$^\star$   & 12.42 & \textbf{66.02} &  & C 256$^\star$   & 12.71 & \textbf{89.86} \\
 & Linear   & 14.77 & 54.75 &  & Linear  & - & - \\
PiCIE+H~\cite{Cho2021PiCIE} & C 27+100  & 14.40 & 50.0 & PiCIE+H~\cite{Cho2021PiCIE} & C 27+100 & - & - \\
ViCE (low-res) & C 27   & 11.40 & 28.91 & ViCE (low-res) & C 27 &  12.81 & 31.87 \\
 & C 27$^\star$   & 11.55 & 50.49 & & C 27$^\star$   & 19.52 & 80.34 \\
 & C 128$^\star$   & 16.66 & 52.33 &  & C 128$^\star$   & 21.48 & 81.55 \\
 & C 256$^\star$   & 17.98 & 54.92 &  & C 256$^\star$   & 21.24 & 81.72 \\
 & Linear   & 25.49 & 62.78 &  & Linear  & \textbf{31.55} & 86.33 \\
 & & & & No pretrain & Linear  & 24.84 & 82.99 \\
 ViCE (high-res) & C 256$^\star$   & \textbf{21.77} & 64.75 & ViCE (high-res) & C 256$^\star$ &  \textbf{25.23} & 84.28 \\
 & Linear   & \textbf{29.38} & \textbf{68.16} &  & Linear  & 30.40 & \textbf{87.0} \\
STEGO$^\ast$~\cite{Hamilton2022STEGO} & C 27   & 28.20 & 56.90 & STEGO$^\ast$~\cite{Hamilton2022STEGO} & C 27   & 21.00 & 73.20 \\
 & Linear   & 41.00 & 76.10 &  & Linear  & - & - \\
\hline\noalign{\smallskip}
\label{tab:low-res_results}
\end{tabular}
\end{center}
\end{table}
\setlength{\tabcolsep}{1.4pt}

\section{Experiments}
\label{sec:experiments}

We implement ViCE in the self-supervised learning framework VISSL~\cite{goyal2021vissl} based on PyTorch~\cite{Paszke2019Pytorch}. The quality of learned embeddings are evaluated on the COCO-Stuff164k~\cite{Lin2014COCO, Caesar2018COCOStuffTA} reduced to 27 classes~\cite{Ji2019IIC} and the Cityscapes~\cite{Cordts2016Cityscapes} benchmark datasets. We use the framework MMSegmentation~\cite{mmseg2020} for evaluation and visualization. Our comparative baseline for dense representation learning is the SOTA unsupervised semantic segmentation CNN model PiCIE~\cite{Cho2021PiCIE} based on DeepCluster~\cite{Caron2018DeepCluster}. We experiment with ResNet 18 and 50 backbones~\cite{He2016ResNet50} and two decoder architectures; the SOTA model DeepLabV3+ (DLV3+)~\cite{Chen2018DeepLabV3Plus} for high-resolution images, and the Feature Pyramid Network (FPN)~\cite{Lin2017FPN} used in our baseline.

We evaluate the semantic richness and spatial accuracy of the resulting embedding maps using clustering and linear models. For unsupervised semantic segmentation we compute a set of $K$ clusters based on output embeddings using FAISS~\cite{Johnson2019FAISS}. Each cluster is greedily assigned the majority label class, or optimally assigned by the Hungarian matching algorithm~\cite{Kuhn1955Hungarian} to cover all classes. For linear model evaluation, we train a $1 \times 1$ convolution layer without a nonlinear activation function. All models are trained and evaluated on separate train and validation sets. Note that the visual concepts learned by ViCE during training are not used for evaluation, and it is therefore fair to compare ViCE and baseline performance as long as the number of clusters is the same in both evaluation models. 

We conduct experiments on 32 V100 32 GB GPUs. Each GPU loads four images, and generates five augmented views. High- and low-resolution views correspond to 512 $\times$ 512 pixels and 256 $\times$ 256 pixels, respectively. The resulting total batch size is 128 images with 640 views. To generating superpixels, we use SLIC~\cite{Achanta2010SLICSuperpixels} implemented in OpenCV~\cite{Bradski2000OpenCV} with average region size 20 px. Maximal mask coverage is 25 \%. The view resize coefficients $\beta$ are sampled between 0.5 to 2. The embedding dimension $D$ and the number of visual concepts $C$ are 128. We use the same set of hyperparameters in all experiments. A hyperparameter study is given in the Appendix.
Parameters for the objective $\mathcal{L}_{cl}$ are the same as SwAV~\cite{Caron2020SwAV}. The FIFO queue consists of 5K score vectors $s^*$ per GPU. The model is optimized using the LARS optimizer~\cite{You2017LARS} with weight decay 10$^{-6}$. The learning rate (LR) schedule is linear warmup followed by cosine decay~\cite{Loshchilov2017SGDRSG, Misra2020PretextInvRepr}. We set the peak LR using the linear LR scaling rule~\cite{Goyal2017ImageNet1H} with a base LR 0.04 for a single 4 GPU node. We initialize models with the default PyTorch pretrained weights obtained by training on ImageNet~\cite{Deng2009ImageNet} for 600 epochs. However, our method can learn from random initialization as shown in Table~\ref{tab:low-res_results}. Timing information is given in the Appendix.

\setlength{\tabcolsep}{4pt}
\begin{table}
\begin{center}
\caption{Performance of best models trained on high- and low-resolution images}
\begin{tabular}{lllll}
\hline\noalign{\smallskip}
Dataset & Resolution & Configuration & Cluster mIoU & Linear mIoU \\
\noalign{\smallskip}\hline
COCO & Low  & RN50, FPN & 19.37 & 27.63 \\
     & High & RN50, DLV3+ & \textbf{21.77} & \textbf{29.38} \\ \noalign{\smallskip}\hline\noalign{\smallskip}
Cityscapes & Low  & RN18, FPN & 21.48 & \textbf{31.55} \\
           & High & RN18, DLV3+ & \textbf{25.23} & 30.40 \\
\hline
\label{tab:high_low_res}
\end{tabular}
\end{center}
\end{table}
\setlength{\tabcolsep}{1.4pt}

\subsection{Representation quality experiments}

\begin{figure}
\begin{center}
\includegraphics[width=1.0\textwidth]{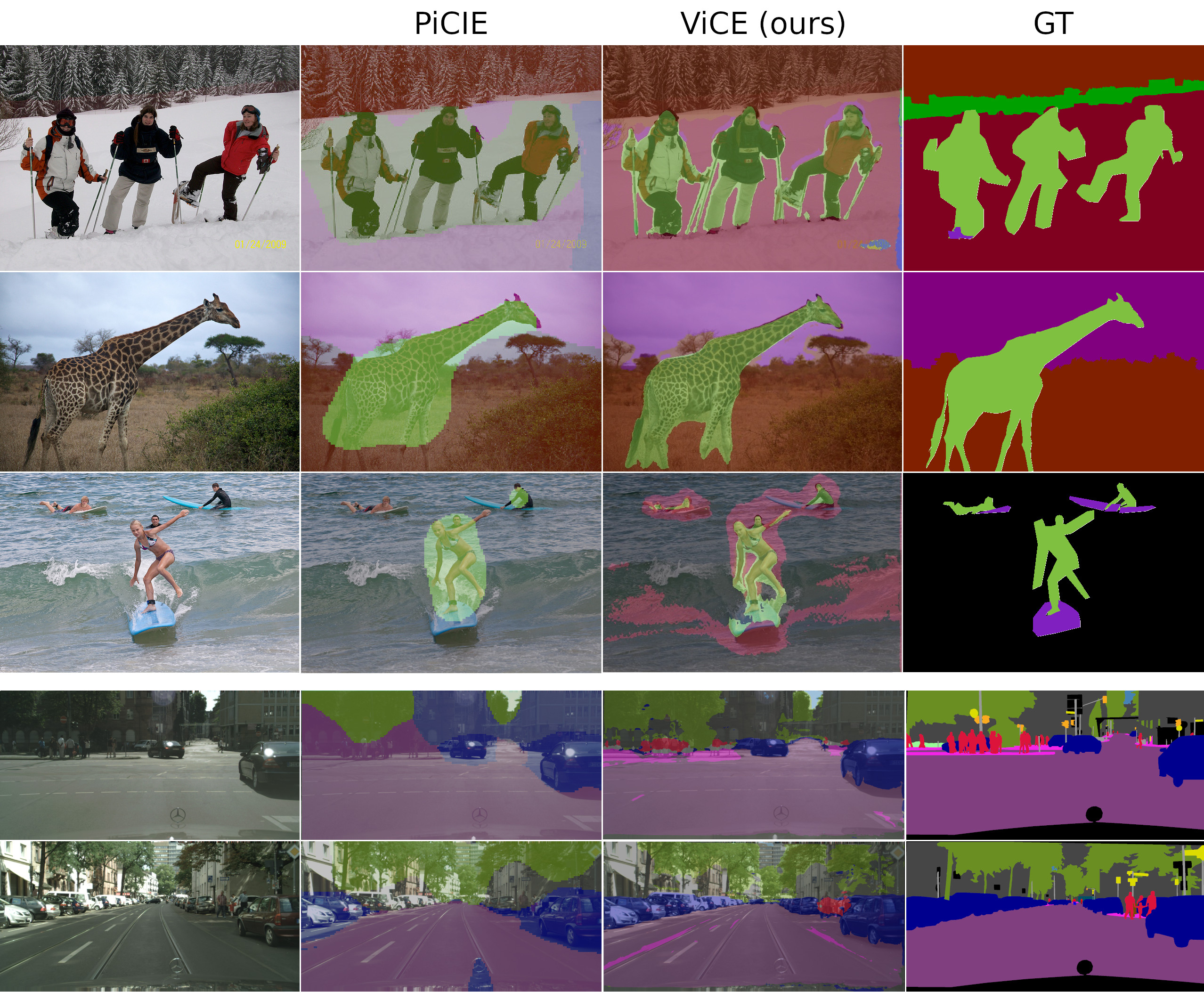}
\end{center}
   \caption{Output cluster visualizations on COCO (top) and Cityscapes (bottom).}
\label{fig:viz_unsup_clustering}
\end{figure}

Table~\ref{tab:low-res_results} presents results on low-resolution image experiments.  C \textit{K} denotes evaluation with \textit{K} clusters, $\diamond$ denotes reproduced results with optimal cluster assignment, $\star$~denotes greedy assignment, and $\ast$ denotes ViT-based models. The best CNN-based cluster and linear model results are written in bold. Both ViCE (low-res) and PiCIE~\cite{Cho2021PiCIE} use the same ResNet 18 backbone, FPN decoder, and $320 \times 320$ px image downsampling procedure for fair comparison. All ViCE models are trained for 4 epochs for COCO, and 24 epochs for Cityscapes, respectively. We trained and evaluated our PiCIE models using the official code~\cite{Cho2021PiCIE}.
Our high-resolution and overclustered model achieves SOTA results on Cityscapes, and on COCO for convolutional models. 
The generic image COCO results show that ViCE is adept at discovering concepts using overclustering~\cite{VanGansbeke2020LearningTC}. We believe this property stems from online clustering being more stable than offline clustering methods~\cite{Caron2020SwAV, Zhan2020OnlineDC}. The Cityscapes results show ViCE improving on PiCIE in all experiments. ViCE performs better than the SOTA ViT-based model STEGO~\cite{Hamilton2022STEGO} on Cityscapes with high-resolution and overclustering.
We trained our best high-resolution C 256* COCO model in 64 h and the equivalent PiCIE model in 52 h.
Fig.~\ref{fig:front_cover},~\ref{fig:viz_unsup_clustering} shows clustering output visualizations.
Table~\ref{tab:high_low_res} shows that the best high-resolution models improves on the best low-resolution models evaluated on high-resolution images. Note that effectively training on high-resolution images is made possible by superpixelization. Results for varying superpixel sizes and performance are given in the Appendix.

\subsection{Ablation studies}

The upper section of Table~\ref{tab:ablation_study} provides an ablation study for low-resolution images evaluated by a linear model. The first column represents the baseline ViCE model using an RN18 backbone and FPN decoder~\cite{Lin2017FPN} without region decomposition. The second columns indicate gains from random masking. The third and fourth column shows gains from applying grid and superpixel region decomposition. The final column indicates that utilizing the more complex DLV3+ decoder~\cite{Chen2018DeepLabV3Plus} is detrimental in the case of low-resolution images. We speculate this is because atrous convolutions in high-resolution decoders skip relevant neighboring information in tiny feature maps. 
The first column in the bottom section of Table~\ref{tab:ablation_study} is empty, as learning dense embeddings for high-resolution images without superpixelization is computationally intractable. The second column showcase the radical difference in using superpixelization. The third column demonstrates the importance of utilizing a high-resolution decoder. The final column shows how superpixels are better than grids with equivalent base element sizes.

\subsection{Domain Generalization experiment}

In Table~\ref{tab:domain_generalization_exp} we show how ViCE benefits when learning from a large general visual domain. Training on COCO and evaluating on Cityscapes with a linear model increases performance from 30.40 to 34.14 (+3.74) mIoU by improving the distinctiveness of complex classes like ``Traffic sign''. Our findings show that general vision models can learn more useful features compared to narrow vision models even when applied in the narrow domain. The recent SOTA model STEGO~\cite{Hamilton2022STEGO} similarly uses a backbone trained on ImageNet only.

\setlength{\tabcolsep}{4pt}
\begin{table}
\begin{center}
\caption{Representation quality ablation study on low- and high-resolution images.}
\begin{tabular}{cccccc}
\hline\noalign{\smallskip}
\multicolumn{6}{c}{\textit{Low-resolution Cityscapes}} \\
     & FPN 1px & Masking & Grid 10px & Super 10px & DLV3+\\ \hline
\noalign{\smallskip}
mIoU & 29.66 & 30.42 & 31.30 & \textbf{31.55} & 11.56\\
Time & 34h 4min & 31h 6min & 5h 31min & \textbf{5h 31min} & 5h 37min\\ \hline
\end{tabular}
\begin{tabular}{ccccc}
\noalign{\smallskip}
\multicolumn{5}{c}{\textit{High-resolution Cityscapes}} \\
     & FPN 1px & FPN super 20px & DLV3+ grid 20 px & DLV3+ super 20px\\ \hline
\noalign{\smallskip}
mIoU & - & 8.98 & 25.53 & \textbf{29.38} \\ 
Time & 92h 20min (est.) & 4h 55min & 10h 1min & \textbf{6h 16min}\\ \hline
\label{tab:ablation_study}
\end{tabular}
\end{center}
\end{table}
\setlength{\tabcolsep}{1.4pt}

\setlength{\tabcolsep}{4pt}
\begin{table}
\begin{center}
\caption{Domain generalization performance}
\begin{tabular}{llll}
\hline\noalign{\smallskip}
Training data domain & Evaluation data domain & mIoU & aAcc\\
\noalign{\smallskip}\hline
Cityscapes & Cityscapes & 30.40 & \textbf{87.00}\\
COCO       & Cityscapes & \textbf{34.14} & 86.10\\
\hline
\label{tab:domain_generalization_exp}
\end{tabular}
\end{center}
\end{table}
\setlength{\tabcolsep}{1.4pt}

\subsection{Qualitative evaluation}

\begin{figure}
\begin{center}
\includegraphics[width=1.0\textwidth]{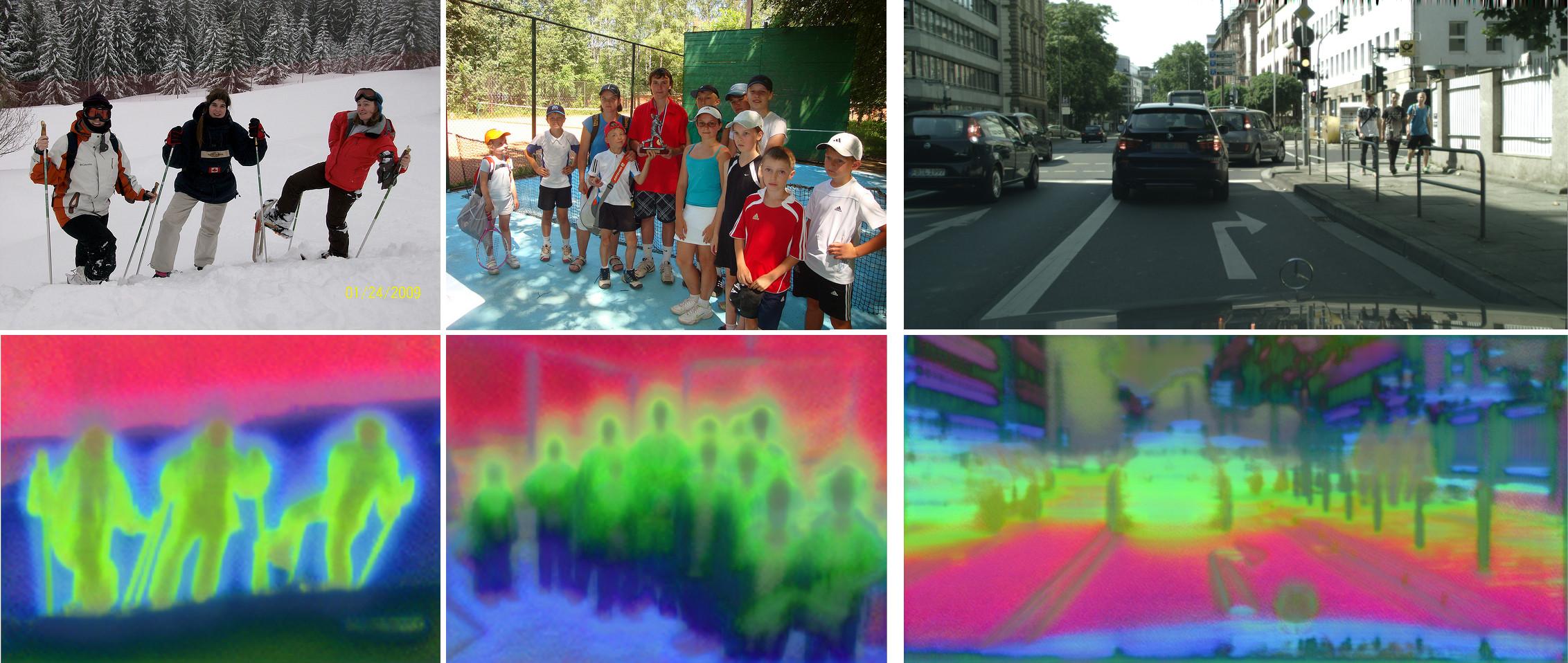}
\end{center}
   \caption{Dense embedding maps visualized as RGB images.}
\label{fig:viz_embs}
\end{figure}

Fig.~\ref{fig:viz_embs} visualizes dense embedding maps to demonstrate how ViCE discovers distinct semantic visual entities or concepts from natural images without human supervision or proposals heuristics~\cite{Uijlings2013SelectiveSearch, Bar2021DETReg}. For example, persons are represented differently from the ground surface, and human faces and bodies are semantically similar. We visualize embedding maps by PCA dimensionality reduction~\cite{Pearson1901PCA} and scale each $z$ to the RGB range.

\section{Conclusion}
\label{sec:conclusion}

We present a new SOTA self-supervised unsupervised semantic segmentation method ViCE for learning to generate dense embedding maps. Our experiments quantitatively demonstrate that decomposing images by superpixelization improves the effectiveness of classification-based self-supervised methods, particularly for high-resolution images, and also achieves better performance than conventional grid decomposition. We hope our work will raise interest in further incorporating non-uniform image decomposition techniques to improve self-supervised computer vision methods including ViT-based models like DINO~\cite{Caron2021DINO} and other dense representation learning methods~\cite{Pinheiro2020VADeR, Wang2012DenseCL, Li2021DSCL, Xie2021PixPro}.

\appendix

\section{Pseudocodes}
\label{sec:sup_a}

Algorithm~\ref{algo:view_gen} explains the generation of $M$ views for a batch of $N$ images. The algorithm samples an image $X^{(n)}$ and computes a superpixel index map $A^{(n)}$. $M$ views are generated from the sampled image and superpixel index map. Each of these views are randomly masked before being resized to the same pixel dimension. Only mutual regions existing in all views are kept. All views are geometrically augmented by random horizontal flipping, and appearance augmented by color distortion and randomly blurred. All generated views are gathered and converted into a 4D tensor.

\begin{algorithm}
\caption{View generation}
\label{algo:view_gen}
\begin{algorithmic}
\State $\tilde{X} := \{\}$ \Comment{Empty sets}
\State $\tilde{A} := \{\}$
\For{$n \in \{1, \dots, N\}$}
    \State $X^{(n)} \sim \text{dataloader}$ \Comment{Sample an image}
    \State $A^{(n)} := \text{superpixels}(X^{(n)})$
    \State $ $
    \State $\tilde{X}^{(n)}, \tilde{A}^{(n)} := \text{gen\_views}(X^{(n)}, A^{(n)})$
    \State $\text{\# } \tilde{X}^{(n)} = \{ \tilde{X}^{(1,n)}, \dots, \tilde{X}^{(M,n)} \}$
    \State $\text{\# } \tilde{A}^{(n)} = \{ \tilde{A}^{(1,n)}, \dots, \tilde{A}^{(M,n)} \}$
    \State $ $
    \State $\tilde{X}^{(n)}, \tilde{A}^{(n)} := \text{mask\_views}(\tilde{X}^{(n)}, \tilde{A}^{(n)})$
    \State $\tilde{X}^{(n)}, \tilde{A}^{(n)} := \text{resize\_views}(\tilde{X}^{(n)}, \tilde{A}^{(n)})$
    \State $\tilde{X}^{(n)}, \tilde{A}^{(n)} := \text{mutual\_regions}(\tilde{X}^{(n)}, \tilde{A}^{(n)})$
    \State $ $
    \State $\tilde{X}^{(n)}, \tilde{A}^{(n)} := \text{geometric\_aug}(\tilde{X}^{(n)}, \tilde{A}^{(n)})$
    \State $\tilde{X}^{(n)} := \text{appearance\_aug}(\tilde{X}^{(n)})$
    \State $ $
    \State $\tilde{X} := \tilde{X} + \tilde{X}^{(n)}$ \Comment{Add new views to set}
    \State $\tilde{A} := \tilde{A} + \tilde{A}^{(n)}$
\EndFor
\State $\tilde{X} := \text{to\_tensor}(\tilde{X})$ \Comment{$\tilde{X} \in \mathbb{R}^{B \times 3 \times h \times w}$}
\State $\tilde{A} := \text{to\_tensor}(\tilde{A})$ \Comment{$\tilde{A} \in \mathbb{R}^{B \times 1 \times h \times w}$}
\end{algorithmic}
\end{algorithm}

Algorithm~\ref{algo:learning_algo} explains the learning algorithm. The model $f_{\theta}$ generates an embedding map $\hat{Z}$ from the image view tensor $\tilde{X}$. The single tensor $\hat{Z}$ is decomposed into $B$ tensors $\hat{Z}^{(b)}$ each corresponding to a single view. Next, four trees are created to contain the latent visual embeddings $z$ for all elements in each mutual region $i$. A mean vectors $z^*$ is computed to represent regions. Each mean vector gets computed a concept compatibility score $s^*$ as distance to each cluster $C = (c^{(1)}, \dots , c^{(K)})$. The swapped prediction objective is computed using the score vectors $s^*$ stored in the tree $T_{S^*}$. The model parameters $\theta$ and set of visual concept vectors $C$ are optimized to reduce the loss $\mathcal{L}$.

\begin{algorithm}
\caption{Learning algorithm}
\label{algo:learning_algo}
\begin{algorithmic}
\State $\text{\# Generate embedding maps}$
\State $\hat{Z} := f_{\theta}(\tilde{X})$ \Comment{$\hat{Z} \in \mathbb{R}^{B \times D \times h \times w}$}
\State $\{\hat{Z}^{(1)}, \dots, \hat{Z}^{(B)}\} := \text{decompose}(\hat{Z})$
\State $ $
\State $\text{\# Create embedding and score trees}$
\State $T_Z(n,m,i) := \{\}$ \Comment{Empty depth-3 trees}
\State $T_{Z*}(n,m,i) := \{\}$
\State $T_{S*}(n,m,i) := \{\}$
\For{$b \in \{1, \dots, B\}$}
    \State $\tilde{Z}^{(b)} := \text{unroll}(\hat{Z}^{(b)})$ \Comment{$\tilde{Z}^{(b)} \in \mathbb{R}^{hw \times D}$}
    \State $\tilde{A}^{(b)} := \text{unroll}(\tilde{A}^{(b)})$ \Comment{$\tilde{A}^{(b)} \in \mathbb{R}^{hw}$}
    \State $n, m := \text{img\_view\_index}(b)$
    \State $I := \text{num\_regions}(\tilde{A}^{(b)})$
    \For{$i \in \{1, \dots, I\}$}
        \State $\text{\# Compute mean vectors for region}$
        \State $\{\hat{z}^{(j)}\} := \text{extract\_region}(\tilde{Z}^{(b)}, \tilde{A}^{(b)}, i)$
        \State $T_Z(n,m,i) := \{\hat{z}^{(j)}\}$
        \State $z^{(i)*} := \text{mean}(T_Z(n,m,i))$
        \State $T_{Z*}(n,m,i) := z^{(i)*}$
        \State $ $
        \State $\text{\# Compute score vectors for region}$
        \State $s^{(i)*} = (T_{Z*}(n,m,i))^T C $
        \State $T_{S^*} := s^{(i)*}$
    \EndFor
\EndFor
\State $ $
\State $\mathcal{L} = \text{swapped\_prediction}(T_{S^*})$
\State $ $
\State $\text{optimize}({\theta, C, \mathcal{L}})$
\end{algorithmic}
\end{algorithm}

The swapped prediction objective is explained in Algorithm~\ref{algo:clustering_obj}. First, we compute an optimal assignment of visual concepts $Q$ based on the scores in the first view $m = 1$. The loss is minimized when predicted visual embeddings in secondary views $m \ge 1$ are closer to the optimally assigned visual concept vectors for each region $i$ in all views $m$ of all images $n$. This results in a cross-entropy optimization objective when both assignments $q^{(i)}$ and compatibility scores $s^{(i)*}$ are normalized.

\begin{algorithm}
\caption{Swapped prediction objective}
\label{algo:clustering_obj}
\begin{algorithmic}
\State $\mathcal{L} := 0$
\State $Q := \text{optimal\_assignment}(T_{S^*})$
\For{$n \in \{1, \dots, N\}$}
    \For{$m \in \{2, \dots, M\}$}
        \For{$i \in \{1, \dots, I\}$}
            \State $q^{(i)} := Q(n,i)$
            \State $s^{(i)*} := T_{S^*}(n,m,i)$
            \State $p^{(i)} := \: \sigma \left( \tfrac{1}{\tau} s^{(i)*} \right)$
            \State $\mathcal{L} \mathrel{-}= q^{(i)} log \: p^{(i)}$
        \EndFor
        \State $\mathcal{L} := \mathcal{L} / I$
    \EndFor
\EndFor
\State $\mathcal{L} := \mathcal{L} / (N (M-1))$
\end{algorithmic}
\end{algorithm}

\section{Hyperparameter study}

We quantify the effect of hyperparameter choices by running a set of high-resolution COCO representation quality experiments for four epochs and linear model evaluation. In each experiment we change only a single parameter in an otherwise static baseline configuration. The experiments are listed in Table~\ref{tab:hyperparam_exp}. Our baseline experiment setup is as follows; view size 512 px, maximal mask coverage 50 $\%$, 128 concepts, queue size of 5K vectors, five views, embedding dimension $D$ equaling 64, and modest view resize range $(0.5, 1.5)$.

The results indicate that modest masking proves to be better than no masking. The ideal number of concepts needs to be found by experiments. Increasing the number of views improves representation learning, as also noted in SwAV~\cite{Caron2020SwAV}. However not by a substantial amount itself explaining the performance gap between ViCE and PiCIE~\cite{Cho2021PiCIE} experiments using five and two views, respectively. Larger embedding size $D$ results in more expressive embeddings. The benefit of increasing $D$ is confirmed by an additional experiment using smaller 400 px view sizes to fit training jobs in GPU memory. All benchmark experiments presented in the main paper use the optimal hyperparameters found in this study.

\setlength{\tabcolsep}{4pt}
\begin{table}
\begin{center}
\caption{Hyperparameter experiments}
\begin{tabular}{ll}
\hline\noalign{\smallskip}
Hyperparameter change & $\Delta$ mIoU\\ \hline
\noalign{\smallskip}
Masking ratio 50\% $\rightarrow$ 25\%                   & +1.52 (+8.3\%)                   \\
Masking ratio 50\% $\rightarrow$ 0\%                    & +1.35 (+7.4\%)                   \\
\#Concepts 128 $\rightarrow$ 64                           & -0.45 (-2.5\%)                   \\
\#Concepts 128 $\rightarrow$ 256                          & -0.59 (-3.2\%)                   \\
Queue size 5K $\rightarrow$ 10K                             & -0.73 (-4.0\%)                   \\
\#Views 5 $\rightarrow$ 2                                 & -1.22 (-6.6\%)                   \\
Emb. size $D$ 64 $\rightarrow$ 32                                 & -1.48 (-8.1\%)                   \\
Resize range (0.5, 1.5) $\rightarrow$ (0.15, 2.0) & -1.86 (-10.1\%)  \\ \hline
\label{tab:hyperparam_exp}
\end{tabular}
\end{center}
\end{table}
\setlength{\tabcolsep}{1.4pt}

Table~\ref{tab:varying_feat_proto} present COCO experiments with varying feature dimension $D$ and number of prototypes $K$. Each model uses the same RN 50 backbone and is trained for 4 epochs. Increasing $D$ consistently results in better performance. However, increasing $K$ beyond 128 prototypes leads to worse results, at least for the same amount of training iterations. The possibility of further improving maximum performance by increasing $D$ and $K$ with additional training epochs remain to be explored.

\setlength{\tabcolsep}{4pt}
\begin{table}
\begin{center}
\caption{Effect of varying feature dimension $D$ and prototype count $K$}
\begin{tabular}{llllllll}
\hline\noalign{\smallskip}
$(D, K)$ & $(64, 64)$ & $(64, 128)$ & $(128, 128)$ & $(128, 256)$ & $(256, 128)$ & $(256, 256)$\\ \hline
\noalign{\smallskip}
mIoU & 26.34 & 26.91 & 27.20 & 26.36 & \textbf{27.25} & 26.08 \\ \hline
\label{tab:varying_feat_proto}
\end{tabular}
\end{center}
\end{table}
\setlength{\tabcolsep}{1.4pt}

\section{Superpixel vs. grid experiments}

\begin{figure}
\centering
\includegraphics[width=\linewidth]{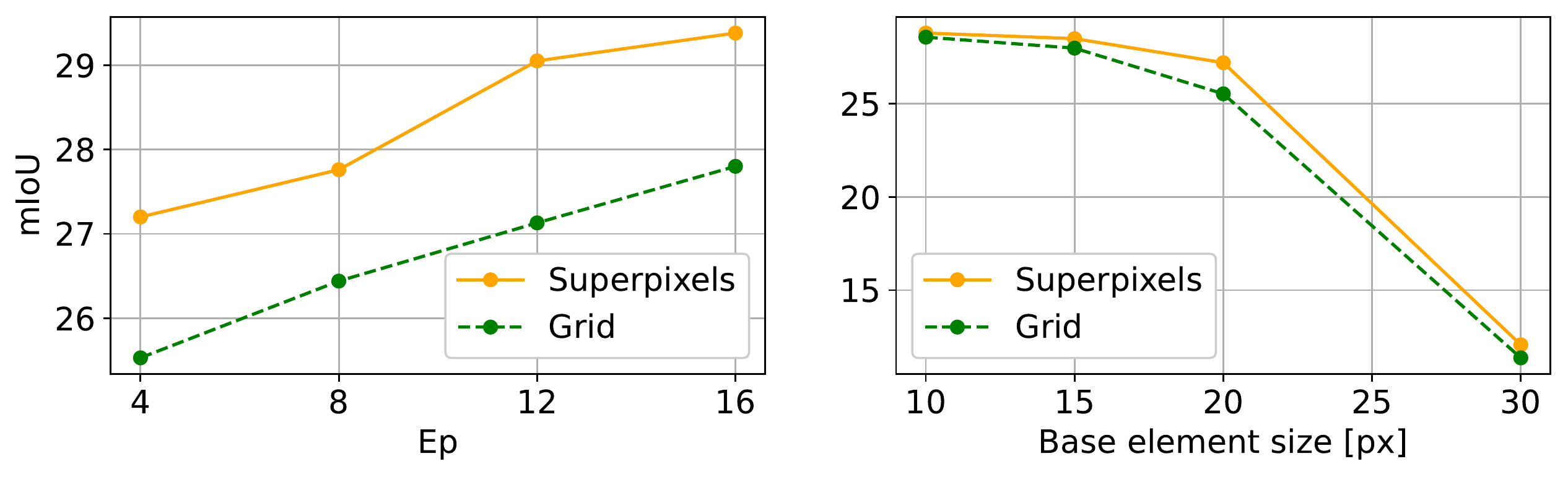}
\caption{Superpixel and grid performance compared on high-resolution COCO}
\label{fig:super_vs_grid}
\end{figure}

The left plot in Fig.~\ref{fig:super_vs_grid} demonstrates consistent gains from using superpixels instead of grids. The right plot shows how performance converges for very small and large base element sizes with linear model evaluation. The result indicates that there exists a sweet spot for base element size in terms of effective learning.

\section{Representation learning from random initialization}

\begin{figure}
\centering
\includegraphics[width=\linewidth]{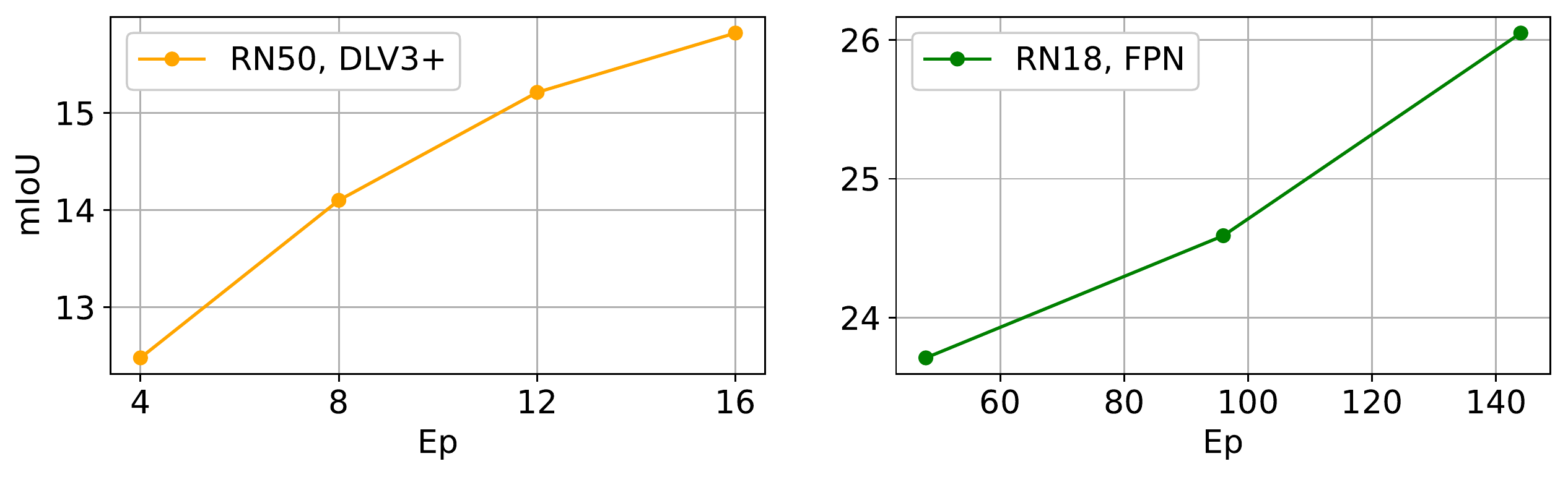}
\caption{Performance when starting from random initialization on high-resolution COCO (left) and low-resolution Cityscapes (right) images}
\label{fig:from_scratch}
\end{figure}

In Fig.~\ref{fig:from_scratch} we show that ViCE is capable to learn visual concepts from scratch using both high- and low-resolution images and linear model evaluation. In particular, the low-resolution Cityscapes model shows linear improvement and achieves 26.05 mIoU after 144 epochs, approaching the best result 30.84 mIoU obtained after 24 epochs starting with pretrained weights. Thus differently from STEGO~\cite{Hamilton2022STEGO}, our method is thus not fundamentally reliant on weight initialization from other supervised or self-supervised pretraining tasks, though using pretrained weights effectively bootstraps learning.

\section{Effect of backbone complexity}

\begin{figure}
\centering
\includegraphics[width=\linewidth]{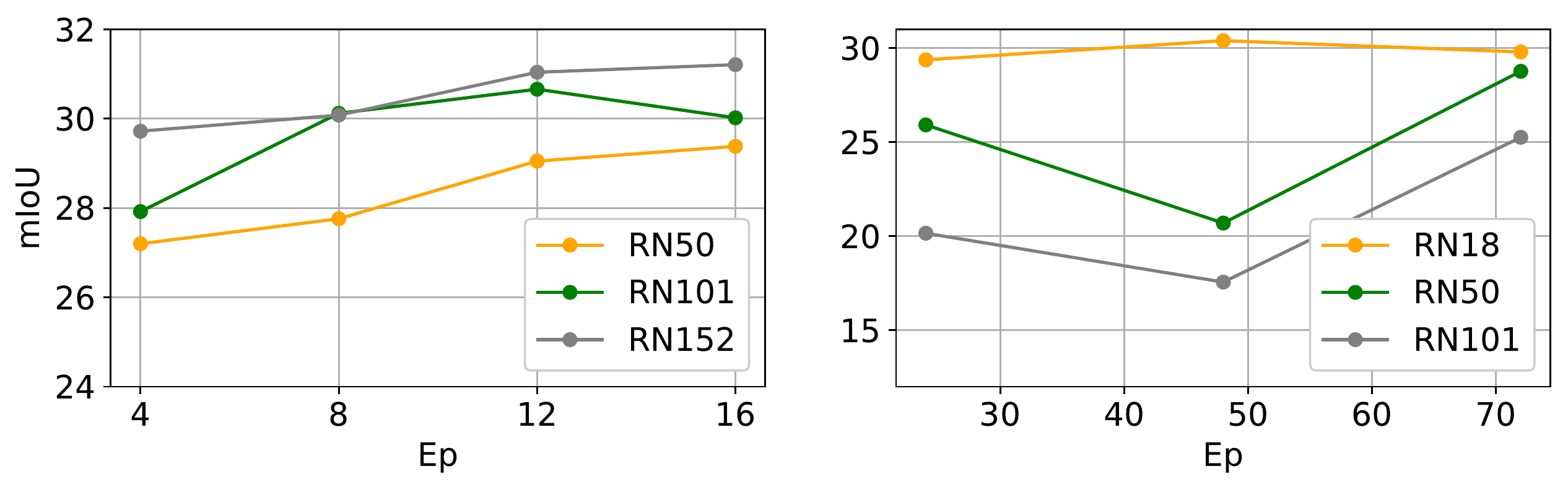}
\caption{Performance with different backbones on high-resolution COCO (left) and Cityscapes (right) images}
\label{fig:backbones}
\end{figure}

In Fig.~\ref{fig:backbones} we show how performance changes with increasing backbone complexity with linear model evaluation. Our results on COCO indicate that performance per epoch consistently improves with increased backbone complexity. In contrast, the results on Cityscapes indicate worse performance. A plausible explanation is that Cityscapes is smaller and less general than COCO, making larger self-supervised models prone to overfit patterns that do not generalize beyond the training sample distribution.

\section{Timing information}

We present training step timing information in Table~\ref{tab:training_step_timings}. The summary is compiled by the framework VISSL~\cite{goyal2021vissl}, and represents average values for a training process involving 32 V100 GPUs distributed over 8 nodes. Table~\ref{tab:inference_step_timings} shows the average inference time per image for cluster and linear evaluation models using a single 3080Ti GPU in a desktop machine. Note that for high-resolution images, linear model evaluation is 160 times quicker than the k-NN cluster evaluation implemented using FAISS~\cite{Johnson2019FAISS}.

\setlength{\tabcolsep}{4pt}
\begin{table}
\begin{center}
\caption{Average training step time per high-resolution image batch}
\begin{tabular}{llllll}
\hline\noalign{\smallskip}
 Phase & Forward & Loss comp. & Backward & Optimization & Tot. \\
 \noalign{\smallskip}
 \hline
 [msec] & 429 & 166 & 4167 & 43 & 4824 \\ 
 \hline
\label{tab:training_step_timings}
\end{tabular}
\end{center}
\end{table}
\setlength{\tabcolsep}{1.4pt}

\setlength{\tabcolsep}{4pt}
\begin{table}
\begin{center}
\caption{Average inference time for a high-resolution image}
\begin{tabular}{llll}
\hline\noalign{\smallskip}
  & Segmentation model & Cluster model & Linear model \\
 \noalign{\smallskip}
 \hline
 [msec] & 57 & 2395 & 15 \\ 
 \hline
\label{tab:inference_step_timings}
\end{tabular}
\end{center}
\end{table}
\setlength{\tabcolsep}{1.4pt}

\section{Additional visualization results}

\begin{figure}
\begin{center}
\includegraphics[width=1.0\textwidth]{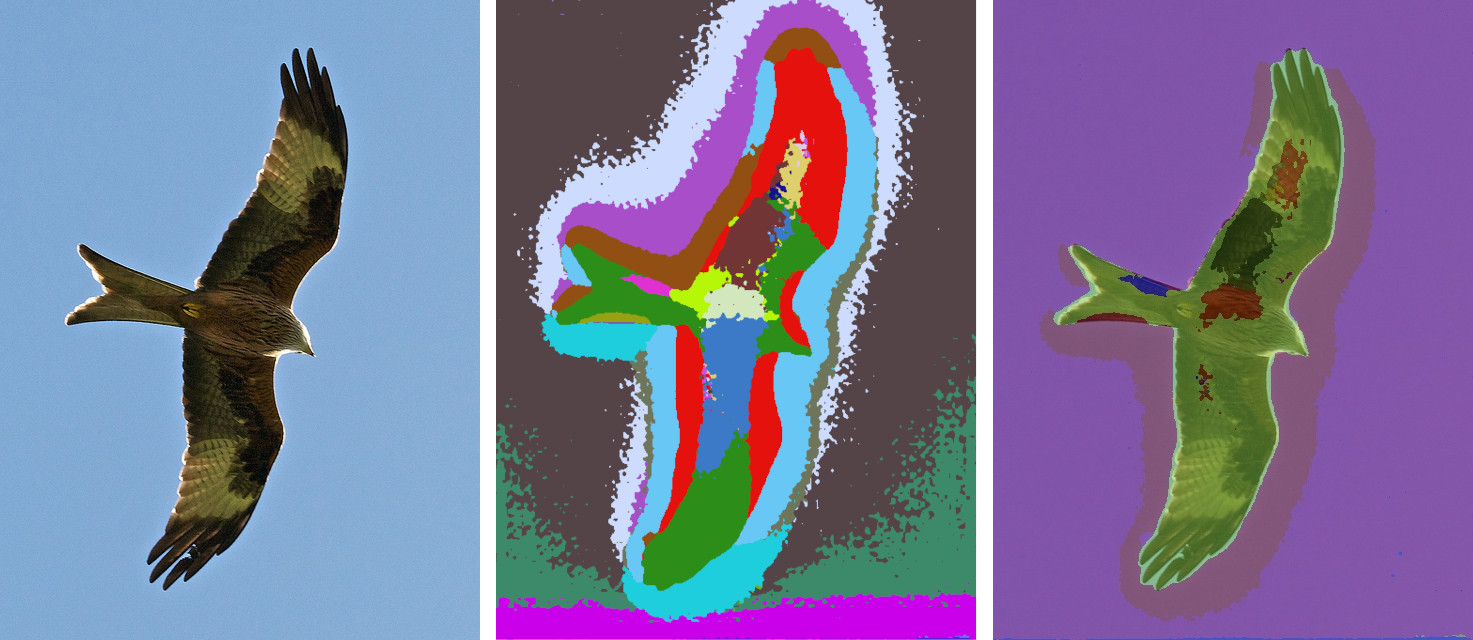}
\end{center}
   \caption{Visualization of output clustering. The center image shows clusters with random colors. The right image shows how clusters are mapped to semantic classes.}
\label{fig:cluster_viz}
\end{figure}

In Fig.~\ref{fig:cluster_viz}, the center image shows how visual concept embeddings in the output embedding map can be clustered into coherent regions. The right image demonstrates how to semantically interpret the image by assigning each cluster a semantic meaning or class. The fact that this is possible depends on the consistent semantic interpretability of the discovered clusters over different samples.

\begin{figure}
\begin{center}
\includegraphics[width=1.0\textwidth]{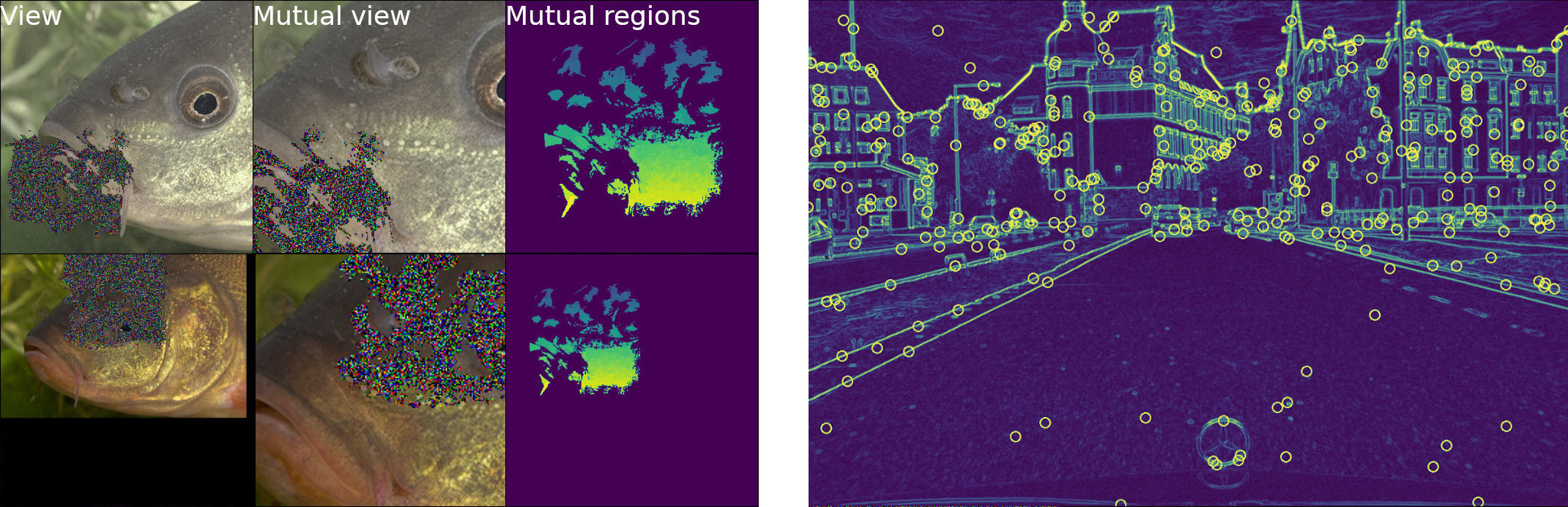}
\end{center}
   \caption{(Left) Examples of two generated view pairs. The first image displays the actual view feed to the model. The second image illustrates the mutual image region. The third image shows mutual superpixel regions colored by region index. (Right) View generation centers sampled from a probability mask representing image complexity measured by the Canny edge detection algorithm \cite{Canny1986CannyEdge}.}
\label{fig:view_generation}
\end{figure}

\begin{figure}
\begin{center}
\includegraphics[width=1.0\textwidth]{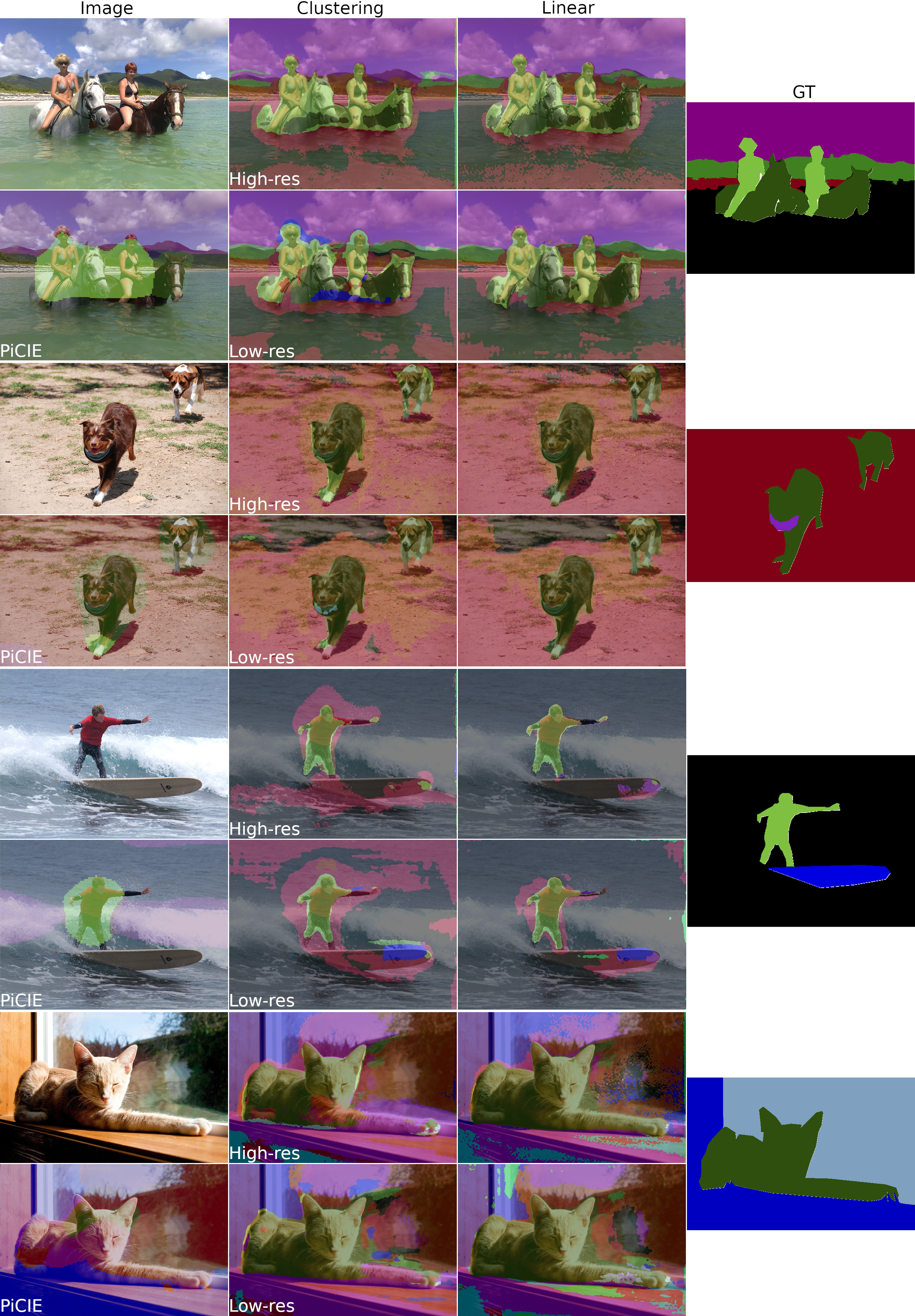}
\end{center}
   \caption{Output visualizations of cluster and linear evaluation models trained on low-
and high-resolution COCO images.}
\label{fig:supplementary_viz}
\end{figure}

Fig.~\ref{fig:supplementary_viz} presents additional output visualizations of high-resolution COCO images for clustering and linear evaluation models with 256 clusters or linear model predictions. Each image is interpreted by five different models and arranged in groups. Each group displays the input image in the top-left corner with the PiCIE output visualization below for comparison. The remaining visualizations display the output of clustering and linear evaluation models trained on high- and low-resolution COCO images. Ground truth labels are visualized in the right column. We find that high-resolution models produce better segmentation borders and less noise. Linear evaluation model output also displays better segmentation borders and less noise, in addition to 160 times faster evaluation time.

\section*{Acknowledgements}

This work was financially supported by JST SPRING, Grant Number JPMJSP2125. The authors would like to take this opportunity to thank the ``Interdisciplinary Frontier Next-Generation Researcher Program of the Tokai Higher Education and Research System''.
\\
\\
The work was financially supported by JSPS KAKENHI, Grant Number 21H04892.
\\
\\
This research was supported by Program on Open Innovation Platform with Enterprises, Research Institute and Academia, Japan Science and Technology Agency (JST, OPERA, JPMJOP1612).
\\
\\
The computation was carried out through the ``General Projects'' program on the supercomputer ``Flow'' at the Information Technology Center, Nagoya University.

\bibliography{egbib}
\end{document}